%% file: main.tex
\documentclass[letterpaper, 10 pt, conference]{ieeeconf}  % Comment this line out if you need a4paper

\IEEEoverridecommandlockouts                              % This command is only needed if 
\input{setup/packages}
\input{setup/acronyms}

\title{\LARGE \bf
Motion Forecasting via Model-Based Risk Minimization}%: \newline Or, How to Ensemble Prediction Models for Autonomous Driving}

\author{Aron Distelzweig$^{1,2}$, Eitan Kosman$^{1}$, Andreas Look$^{1}$, Faris Janjo\v{s}$^{1}$, Denesh K. Manivannan$^{1}$, Abhinav Valada$^{2}$
\thanks{$^{1}$Bosch Center for Artificial Intelligence, Germany, Israel.}%
\thanks{$^{2}$Department of Computer Science, University of Freiburg, Germany.}%
}

\begin{document}
\maketitle
 \thispagestyle{empty}
\pagestyle{empty}

% ----------------------------- Abstract ------------------------------
\input{sections/abstract}

% ----------------------------- Intro ---------------------------------
\input{sections/introduction}

% ----------------------------- Related Work ----------------------------
\input{sections/related_work.tex    }

% ----------------------------- APPROACH ------------------------------
\input{sections/approach}

% ----------------------------- RESULTS ----------------------------
\input{sections/experiments}

% ----------------------------- Conclusion ----------------------------
\input{sections/conclusion}

% ----------------------------- Appendix ------------------------------
%\addtolength{\textheight}{-12cm}   
\clearpage

% ----------------------------- Bib ------------------------------
\bibliographystyle{IEEEtran}
\bibliography{main}

\end{document}

%% file: setup/packages.tex
\usepackage{booktabs}
\usepackage{hyperref}
\usepackage{caption}
\usepackage{multirow}
\usepackage{url}
\usepackage{todonotes}
\usepackage{algorithm}
\usepackage{algpseudocode}
\usepackage{blindtext}
\usepackage{amsmath}
\usepackage{amsfonts}
\usepackage{bm}
\usepackage{cite}
\usepackage{flushend}
\usepackage[nolist]{acronym}
\usepackage[hang,flushmargin]{footmisc}

%% file: setup/acronyms.tex
\begin{acronym}
\acro{ML}{Machine Learning}
\acro{GPU}{Graphics Processing Units}
\acro{DNN}{Deep Neural Networks}
\acro{CNN}{Convolutional Neural Network}
\acro{RNN}{Recurrent Neural Networks}
\acro{GNN}{Graph Neural Networks}
\acro{GAT}{Graph Attention Network}
\acro{GCN}{Graph Convolutional Networks}
\acro{GRU}{Gated Recurrent Unit}
\acro{LSTM}{Long Short-Term Memory}
\acro{KF}{Kalman Filters}
\acro{BEV}{Bird's Eye View}
\acro{GAN}{Generative Adversarial Network}
\acro{MTP}{Multi-Modal Trajectory Prediction}
\acro{MLP}{Multilayer Perceptron}
\acro{MAE}{Mean Absolute Error}
\acro{MSE}{Mean Squared Error}
\acro{ADE}{Average Displacement Error}
\acro{FDE}{Final Displacement Error}
\acro{minADE}{Minimum Average Displacement Error}
\acro{minFDE}{Minimum Final Displacement Error}
\acro{VAE}{Variational Autoencoders}
\acro{CVAE}{Conditional Variational Autoencoder}
\acro{AV}{Automated Vehicle}
\acro{MHA}{Multi-Head Attention}
\acro{AD}{Automated Driving}
\acro{DL}{Deep Learning}
\acro{RL}{Reinforcement Learning}
\acro{MR}{Miss Rate}
\acro{IL}{Imitation Learning}
\acro{MDP}{Markov Decision Process}
\acro{POMDP}{Partially Observable Markov Decision Process}
\acro{RSSM}{Recurrent State Space Models}
\acro{VAE}{Variational Autoencoder}
\acro{UT}{Unscented Transform}
\acro{GMM}{Gaussian Mixture Model}
\acro{NF}{Normalizing Flow}
\acro{KL}{Kullback-Leibler}
\acro{UT}{Unscented Transform}
\acro{FID}{Fréchet Inception Distance}
\acro{NLL}{Negative Log Likelihood}
\acro{ELBO}{Evidence Lower Bound}
\acro{WTA}{Winner-Takes-All}
\acro{EM}{Expectation Maximization}
\acro{NMS}{Non-Maximum Suppression}
\acro{OOD}{Out of Distribution}
\end{acronym}

%% file: sections/abstract.tex
\begin{abstract}
Forecasting the future trajectories of surrounding agents is crucial for autonomous vehicles to ensure safe, efficient, and comfortable route planning. While model ensembling has improved prediction accuracy in various fields, its application in trajectory prediction is limited due to the multi-modal nature of predictions.
In this paper, we propose a novel sampling method applicable to trajectory prediction based on the predictions of multiple models. We first show that conventional sampling based on predicted probabilities can degrade performance due to missing alignment between models. 
To address this problem, we introduce a new method that generates optimal trajectories from a set of neural networks, framing it as a risk minimization problem with a variable loss function. By using state-of-the-art models as base learners, our approach constructs diverse and effective ensembles for optimal trajectory sampling.
Extensive experiments on the nuScenes prediction dataset demonstrate that our method surpasses current state-of-the-art techniques, achieving top ranks on the leaderboard. We also provide a comprehensive empirical study on ensembling strategies, offering insights into their effectiveness.
Our findings highlight the potential of advanced ensembling techniques in trajectory prediction, significantly improving predictive performance and paving the way for more reliable predicted trajectories.
\end{abstract}

%% file: sections/introduction.tex
\section{Introduction}
\label{sec:introduction}
% \todo{@Faris, Eitan}
% \blindtext[1]

Predicting the future motion of surrounding agents is a crucial component of \ac{AD} systems~\cite{khan2022level}. Future predictions, often in the form of trajectories, are used to facilitate the decision-making of a planner, which generates actions in order to execute a safe, efficient, and comfortable trajectory to the target destination. 
Complex driving scenarios, such as urban environments with rich information, require state-of-the-art \ac{ML} models like \ac{GNN}~\cite{waikhom2023survey, buchner20223d}, Transformers~\cite{vaswani2017attention, gosala2023skyeye}, and various time-series processing techniques~\cite{chen2023long}, to address the problem.
This enables them to account for the temporal aspects of the problem while at the same time considering interactions between the entities within a driving scene. 
Given multiple valid solutions for the future behavior of traffic participants, these architectures often generate several predictions~\cite{Rudenko_2020}. 
Capturing the plurality of future motion, so-called multi-modality, in rich environmental contexts necessitates many predictions. 
Consequently, the set of valid predictions can become exceedingly large and prohibitively difficult to reason about for a planner. Hence, it is essential to obtain a minimal representative subset that trades off the diversity of multiple futures while accounting for their plausibility.

Meanwhile, ensemble learning~\cite{zhou2012ensemble} has proved to be a useful tool to meaningfully provide a set of multiple predictions from learned models. In many fields, simply combining multiple learners, via different instances of the same model or from heterogeneous models, has shown great success in improving prediction accuracy, robustness, and especially diversity of the overall output \cite{pmlr-v139-izmailov21a}. However, their usage in trajectory prediction remains limited due to the multi-modal nature of the output. According to~\cite{gilles2022gohome}, model-ensembling necessitates a strategy for determining which predictions from different learners can be combined, presenting combinatorial challenges. On the one hand, this difficulty renders ensembles an inconvenient tool for the task. On the other hand, the potential benefit opens the door to a promising research direction, aiming to mitigate the gap between trajectory prediction and ensembles.
% \todo[inline]{Detail out why trajectory selection problem is important}

\begin{figure*}[ht!]
\begin{center}
\includegraphics[width=2\columnwidth]{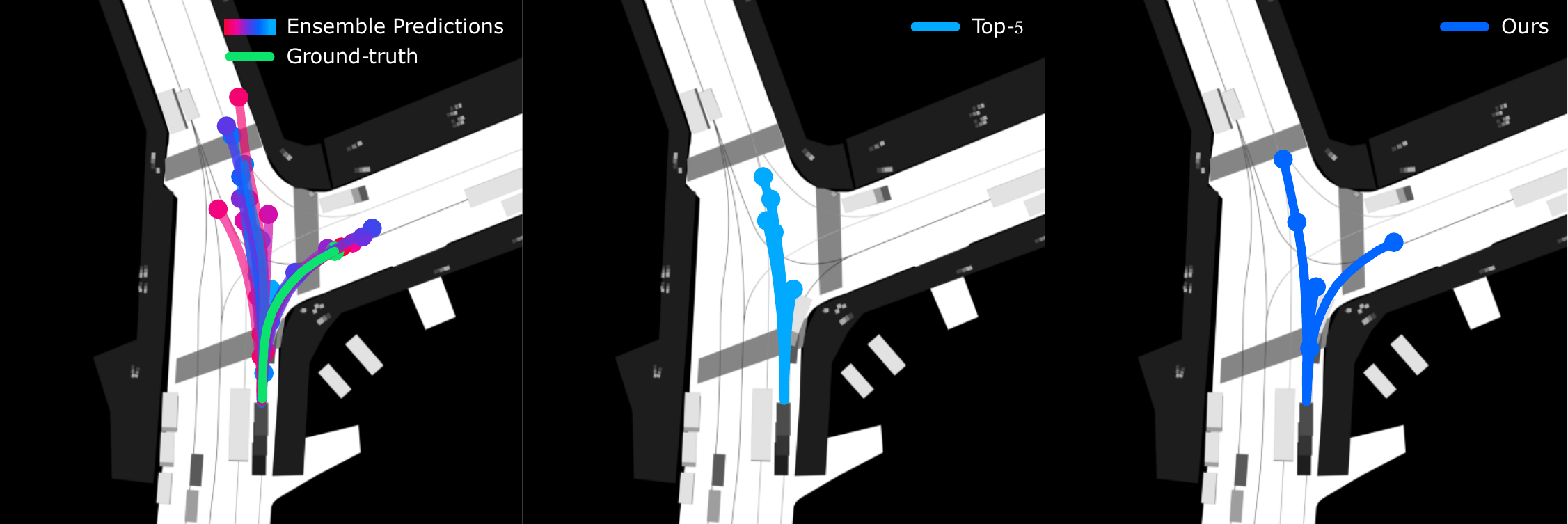}
\end{center}
\caption{Illustration of future trajectories. \textbf{Left:} All 30 proposals from our ensemble, including LaPred~\cite{kim2021_lapred}, LAformer~\cite{liu2024_laformer}, and PGP~\cite{deo2022_pgp}, alongside the ground-truth trajectory. \textbf{Middle:} Five out of 30 trajectories with the highest predicted probabilities. \textbf{Right:} Five out of 30 trajectories generated with our proposed method.}
\label{fig:ours_vs_topk}
\end{figure*}

% \todo[inline]{shorten related work in intro as we have a detailed discussion later on}
% The aforementioned research direction has already spurred several works.
The successful usage of ensembles in motion prediction literature is restricted either to the heatmap prediction models of~\cite{gilles2021home,gilles2022gohome}, whose occupancy representation facilitates combining ensemble component outputs or as a tool within specific architectures such as~\cite{varadarajan2022multipath++, li2022ensemble}, which outputs trajectories but ensembles latent vectors over multiple instances of their model in order to increase output diversity. An alternative approach to employing ensembles is using the output representations of several current state-of-the-art models in leading public leaderboards, such as nuScenes~\cite{nuscenes}. As they provide scores along with predictions, we can combine their strengths by constructing ensembles from these models and sample predictions based on their scores, such as using top-k sampling. However, we demonstrate here that naively sampling based on the highest scores degrades performance. Instead, we propose an alternative method that achieves optimal performance w.r.t. a general definition of optimality. To illustrate this in a real example, Fig.~\ref{fig:ours_vs_topk} shows the sub-optimal selection of the five most probable predictions from a set generated by three different models, in turn inadequately covering the range of possibilities. Conversely, our sampling strategy yields a set of the same size that more comprehensively represents valid possibilities.

The beneficial ensembling methods in trajectory prediction demonstrated by~\cite{gilles2021home} and \cite{li2022ensemble} raise further interest regarding the optimality of their ensembling mechanisms. We argue that employing a more sophisticated sampling strategy, rather than relying on simple averaging or plurality voting as employed in \cite{li2022ensemble} could serve as an alternative mechanism to significantly enhance trajectory prediction performance. This approach has the potential to further optimize the performance of state-of-the-art base learners. To this end, we propose a new sampling method, referred to as the generation of \textit{optimal} trajectories from a set of neural networks. Assuming that the base learners can approximate the true risk, this method can be interpreted as a risk minimization problem, where a chosen loss function represents the risk, and the predicted trajectories form a learner distribution approximating the ground truth. Our approach employs state-of-the-art models such as PGP~\cite{deo2022_pgp}, LAformer~\cite{liu2024_laformer}, and LaPred~\cite{kim2021_lapred} as base learners. This selection ensures the construction of highly performant ensembles from strong and diverse bases, as these models originate from different families. This diversity enhances the ensemble's ability to capture a wider range of patterns, thereby improving overall prediction accuracy. {In summary, we:}

% \todo[inline]{merge 3 + 4}
\begin{enumerate}
    \item Demonstrate that sampling based solely on predicted probabilities degrades performance in ensembles, highlighting the limitations of this naive approach.
    \item Introduce a novel method for sampling trajectories from proposals, showing that it outperforms advanced sampling techniques and enhances predictive accuracy.
    \item Present empirical results demonstrating that our proposed method surpasses state-of-the-art prediction methods on the nuScenes prediction dataset. While surpassing the performance of the base learners by a large margin, we additionally observe our process to be ranked amongst the best-performing methods on the nuScenes leaderboard. Finally, through a more comprehensive empirical study on sampling and ensembling strategies, we provide valuable insights and highlight the strengths and weaknesses of various approaches.
\end{enumerate}

% Proposal: separate contributions of optimal sampling (which can be used with any discrete distribution generator: ensembles, dropout, generative models) and a successful application with ensembled model, which is a successful method outside of trajectory prediction.

%% file: sections/related_work.tex
\section{Related Work}
\label{sec:related_work}
Our proposed approach first ensembles diverse trajectory prediction models to generate sets of weighted trajectories, followed by a resampling step that applies our proposed sampling strategy. 
%This two-step process ensures that the ensemble captures a wide range of possible future trajectories while the resampling step refines the predictions, selecting the most plausible and representative trajectories for use in downstream tasks. 
Thus, we structure this section along with related research topics: approaches to generate {sets of trajectories} along with their probabilities, i.e., a discrete distribution of multi-modal trajectories, sampling trajectories from distributions, and usage of {ensemble techniques} in prediction.

\subsection{Weighted Trajectory Sets}
Many motion prediction approaches in the AD context directly regress a fixed set of trajectories along with their likelihoods, thus representing a categorical distribution of trajectory proposals. 
These trajectories can also represent the means of a parametric mixture distribution with separately regressed covariances. 
To achieve this, existing approaches focus on various aspects of the problem, from inferring relationships to surrounding map elements~\cite{kim2021_lapred,deo2022_pgp,liu2024_laformer,hallgarten2024stay}, modeling interaction between traffic participants~\cite{tolstaya2021identifying,sun2022m2i,luo2023jfp,shi2024mtr++}, to modeling multi-modality of future motion~\cite{zhao2021tnt,varadarajan2022multipath++,shi2022_mtr}. 
In these contexts, various learned models are used to encode rich contextual information and generate trajectory outputs, especially prominent being \ac{GNN}s~\cite{hamilton2020graph} and Transformers~\cite{vaswani2017attention}. Many such approaches are deterministic in nature or model distributions without prior assumptions.
%\todo[inline]{emphasize that these approaches are mostly deterministic? --> yes, one sentence}

%the regressed distributions are not used to perform sampling. Thus, such approaches are not of interest in this work as a direct comparison, however they are repurposed in order to obtain a distribution amenable to sampling.

%% they do sample: generative models / probabilistic parametric
In contrast to approaches that directly regress a fixed set of trajectories, generative models in prediction have the ability to provide an arbitrary number of trajectories by sampling from learned latent distributions. In this context, various \ac{CVAE}~\cite{casas2020implicit,salzmann2020trajectron++,yuan2021agentformer,janjovs2023conditional}, \ac{GAN}~\cite{huang2020diversitygan,dendorfer2021mg,gomez2022exploring}, or diffusion-based~\cite{luo2023jfp,jiang2023motiondiffuser} prediction approaches exist. Common among these classes of models is the flexibility in constructing sets of trajectories, however, providing likelihoods along with trajectories usually does not come 'out of the box'. Various strategies exist for this: the \ac{CVAE} approaches~\cite{casas2020implicit,janjovs2023conditional} learn additional classifier networks, while the diffusion approach in~\cite{jiang2023motiondiffuser} integrates out the diffusion iterations in order to compute the log probability of a sample. Another generative approach is~\cite{seff2023motionlm}, an autoregressive maximum-likelihood approach inspired by language modeling and without prior distribution assumptions, which casts trajectory prediction as the problem of learning a discrete sequence in a continuous domain. In order to obtain a representative set of candidate trajectories,~\cite{seff2023motionlm} uses \ac{NMS} and KMeans -- tools that are shown to be inferior to our proposed approach (see Sec.~\ref{sec:results}).

Our proposed optimal sampling approach is not restricted to a specific methodology to generate a weighted trajectory set. Guided by simplicity and ease-of-use, we opt for the direct approaches found in~\cite{kim2021_lapred,deo2022_pgp,liu2024_laformer}.

\subsection{Sampling Trajectories from Distributions}
The concept of sampling from distributions in order to obtain trajectory candidates arises in various forms: (i) constructing and sampling a parametric continuous distribution, (ii) constructing a non-parametric distribution and drawing samples via discrete optimization, or (iii) sub-sampling a fixed set of trajectories. 

Approaches employing (i) are typically generative models where latent distributions are explicitly modeled and samples are drawn and decoded into trajectories, usually via random sampling~\cite{ivanovic2020multimodal,casas2020implicit}. 
In contrast,~\cite{janjovs2023conditional,look2023_cheap} employ deterministic sampling to draw latent vectors and subsequently trajectories from fixed points in the domain. 

Since we do not construct explicit continuous distributions, our optimal sampling approach is conceptually closer to approaches in (ii). Here, works such as~\cite{gilles2021home,gilles2022gohome,gilles2022thomas} predict future occupancy maps that can be interpreted as probability distributions\footnote{Although the predicted occupancy maps could be considered as a mixture of Beta distributions over a fixed grid, we group these approaches under (ii) since there are no assumptions around the overall shape of the distribution, i.e., the heatmap.}. The occupancies are then sampled in a discrete optimization manner by minimizing the \ac{MR} and \ac{FDE} metrics of drawn trajectories. Our optimal sampling method in~Eq. (\ref{eq:optimal_trajectories}) draws inspiration from this approach, however, our optimization is continuous and differentiable, thus facilitating the use of automatic differentiation libraries. 

We use our optimal sampling procedure to sub-sample from a set of trajectories (iii). In this context, a notable approach is~\cite{gu2021densetnt}, which first constructs a dense set of discrete goals and then applies an optimization step to obtain a representative subset. Optimizing over goals puts strong requirements on preserving diversity for heterogeneous heatmaps; in order to avoid applying and tuning \ac{NMS} (as the predecessor goal-prediction approach in~\cite{zhao2021tnt}), they introduce another, offline-trained goal set prediction task. Our continuous optimization approach is simpler and more flexible since we directly optimize over full trajectories while offloading the task of ensuring diversity to ensemble components. In a similar fashion of discrete optimization,~\cite{varadarajan2022multipath++} generates clusters of decoded trajectories in an \ac{EM} manner, starting from a pre-computed and sufficiently diverse initialization. In comparison, our gradient-based optimization approach directly optimizes the target metrics and does not require a 'good-enough' initialization. Additionally, approaches in~\cite{seff2023motionlm,jiang2023motiondiffuser} use \ac{NMS} or KMeans to sub-sample trajectories; we ablate our sampling strategy against such tools in Sec~\ref{sec:results}. Finally, another class of related approaches are~\cite{phan2020_covernet, schmidt2023reset}, which directly generate a set of trajectories with the help of kinematic motion constraints and thus replace the task of trajectory regression with learned classification from the set. Such classification-based predictors are not well-represented in the literature since they place strong demands on context understanding, thus, our proposed method favors regression-based predictors since they benefit from strong priors of past motion.

%Denesh (ren2021safety,mohajerin2019multi)
% Another set of works \cite{gilles2021home,gilles2022gohome} focus on incorporating multi-modality in the predicted trajectories by predicting occupancy maps that can be interpreted as probability distributions for the agents' trajectories. In particular, they leverage HD-map features to generate a heat-map output representing a  probability distribution for the agent's future trajectories. The trajectories are then sampled by minimizing the Miss-rate (MR) and Final Displacement error (FDE) metrics. Our Optimal sampling approach that samples trajectories by jointly minimizing the Average Displacement error (ADE) draws inspiration from this approach.
%Denesh
%%
%Given a set of candidate trajectories, various sampling techniques are used to refine the search space. For instance, ( Sampling-Based Optimal Trajectory Generation for Autonomous Vehicles Using Reachable Sets , \cite{manzinger2020using} , \cite{althoff2014online}) use Reachability analysis to prune candidate trajectories. However, they don't focus on complicated scenarios with intersections where diversity in predicted modes become important. (HOME, GOHOME , ...)
%%
\subsection{Ensembles in Motion Prediction}

%Combining results from multiple learners to enhance the diversity and performance of trajectory prediction models is not well explored, but offers great potential to enhance model performance. Some works like \cite{min2019rnn,min2020interaction} that regress a fixed trajectory, propose to use deep ensembling  to provide a variance estimate for the regressed trajectory, where trajectory predictions of multiple \ac{DL} models belonging to the same class are utilized. Other works utilizing deep ensembles include \cite{kim2021hybrid}, where a hybrid approach to combine physics-based and \ac{DL} based trajectory prediction models to enhance performance is explored.
Ensembling multiple learned models to enhance the diversity and performance is sparsely explored within trajectory prediction, despite the demonstrated benefits in other fields. Among the first to use ensembles of learned models is~\cite{gilles2022gohome} in the context of heatmap prediction. This output representation allows direct positional comparison of outputs across multiple diverse heatmap generators, combining the \ac{GNN}-based~\cite{gilles2022gohome} and \ac{CNN}-based~\cite{gilles2021home}, thus boosting overall performance. In contrast,~\cite{li2022ensemble} demonstrates the use of ensembles for non-grid representations. Their multi-task architecture simultaneously predicts driving maneuvers as a classification task and vehicle trajectories as a regression task. Both outputs undergo ensembling, using plurality voting for maneuver classification and simple averaging for trajectory regression, effectively combining base learner strengths and improving predictive accuracy and robustness in varied traffic scenarios. 
Similarly,~\cite{varadarajan2022multipath++} ensembles their proposed trajectory prediction model across several instances, boosting its performance by first clustering the ensemble component outputs w.r.t. overall diversity and secondly, an \ac{EM} optimization contingent on a suitable initialization. 
In a similar fashion, the use of dropout can be interpreted as a form of implicit ensembling; it is used in~\cite{nayak2022uncertainty}
%\todo{find better reference than nayak2022uncertainty} 
for boosting overall performance and in~\cite{min2020interaction,janjovs2023bridging} for epistemic uncertainty quantification. 
Overall, common among many approaches in literature is the ensembling of multiple instances of the same model. In contrast, our approach is the first to successfully apply the ensembling of heterogeneous trajectory prediction models in order to surpass their individual performance, as well as the performance of ensembles containing purely those models. 
%Ensembling approaches and dropout are primarily used for epistemic uncertainty quantification~\cite{filos2020can}. 
%Dropout in prediction:~\cite{nayak2022uncertainty,janjovs2023bridging}

%% file: sections/approach.tex
\section{Technical Approach}
\label{sec:approach}

In this section, we introduce our novel method to generate \textit{optimal} trajectories by combining predictions from an ensemble of heterogeneous trajectory prediction models. 
We refer to an ensemble as a set of neural networks.
Our method is independent of how the ensemble is constructed, meaning the ensemble can be generated through various methods such as deep ensembling \cite{Lakshminarayanan17_ensembles}, dropout \cite{gal16_dropout}, SGLD \cite{welling2011_sgld}, or any other technique.
We frame our approach as a risk minimization problem \cite{hastie2009elements}, where the ensemble approximates the true risk.
We observe that traditional methods, which either sample from the distribution over future trajectories or select the most likely trajectories, degrade in performance as more ensemble members are added. 
In contrast, our method improves prediction performance by incorporating additional networks into the neural network ensemble.

We present our novel method in three steps. 
First, we discuss trajectory prediction models, ensembles, and establish the formal basis of our approach.
Second, we define what we refer to as \textit{optimal} trajectories. 
Lastly, we present our method and demonstrate its ease of application in Alg.~\ref{alg:main}.

\subsection{Ensembles of Trajectory Prediction Models}

In this work, we consider neural network-based trajectory prediction models $ f^m(x) \rightarrow \{w_n^m, y_n^m\}_{n=1}^N$ that predict $N \in \mathbb{N}_+$ proposal trajectories for a target agent, as commonly described in concurrent literature \cite{kim2021_lapred, deo2022_pgp, schaefer2023_caspnet, phan2020_covernet}.
Here $y_n^m \in \mathbb{R}^{T\times 2}$ represents the $n$-th proposal trajectory of the $m$-th model and $w_n^m \in \mathbb{R}^{+}$ represents the corresponding weight. 
Each proposed trajectory $y_n^m$ spans $T$ time steps and consists of 2 features: the position from a top-down view.
The weights $w_n^m$ form a standard $N$-simplex. 
The input $x \in \mathbb{R}^{D_x}$ consists of $D_x$ features, typically including past trajectories of both the target agent and surrounding agents, as well as the map.
Additionally, some models do not predict a weighted set of trajectories but instead, use a \ac{GMM} and rely on the mean values of the \ac{GMM} as their proposed trajectories \cite{liu2024_laformer, shi2022_mtr}.

Suppose, we are given an ensemble of \text{$M \in \mathbb{N}_+$} trajectory prediction models. 
We denote the set of all predictions as  $\{w, y\} = \{\{w_n^1, y_n^1\}_{n=1}^N, \ldots, \{w_n^M, y_n^M\}_{n=1}^N \}$, which consists in total of $MN$ trajectory-weight pairs. 
Essentially, this set includes predictions of $M$ models, each providing $N$ weighted trajectories. 
We rewrite the set of all predictions as a distribution
\begin{equation}
    q(y) =  \sum_{m=1}^M\sum_{n=1}^N \frac{w_n^m}{M} \delta(y_n^m - y),
    \label{eq:weighted_delta_distribution}
\end{equation}
which is a weighted sum of Dirac delta distributions that in turn can be interpreted as a categorical distribution. 
Assuming uniform weighting among the models, we divide the weights $w_n^m$ by $M$ to ensure that the double sum over the weights evaluates to $1$.  
We also assume that the true distribution $p(y)$ over future trajectories is sufficiently well modeled by our approximation, i.e., $p(y) \approx q(y)$.

\subsection{What are Optimal Trajectories?}
Having introduced the probabilistic interpretation, our next task is to sample $S \in \mathbb{N}_+$ \textit{optimal} trajectories, denoted as $\hat{y} = \{\hat{y}_s\}_{s=1}^S$. 
We define a set of predictions as \textit{optimal} if it minimizes a risk, which is the expected value of a loss function $\mathcal{L}(y, \hat{y} ) \rightarrow \mathbb{R}_+$. 
This set of optimal predictions is also commonly referred to as the Bayes estimator \cite{berger1985statistical, lehmann2003theory}.
Formally, the optimal set of predictions is determined as the argmin solution to 
\begin{equation}
    \hat{y} = 
    \underset{\tilde{y}}{\operatorname{argmin}} ~
    \mathbb{E}_{y\sim p(y)} \left[ \mathcal{L}(y, \tilde{y})\right].
    \label{eq:definition_optimal}
\end{equation}
Above, $y \sim p(y)$ represents samples from the ground truth distribution over future trajectories $p(y)$.
The concept of risk minimization is widely applied in various domains, such as dynamical system learning \cite{haussmann2021_pacsde} and parameter estimation of neural nets \cite{morningstar22a_pacm}.
Now, our task is to select a suitable loss function $\mathcal{L}$. 
A commonly used metric for evaluating the quality of trajectory prediction models is the $\text{minADE}_k(y, \hat{y})$ \cite{nuscenes}, which computes the minimal average displacement error over $k\in \mathbb{N}_+$ trajectories. 
Moreover, alternative loss functions like collision rate \cite{suo2021_trafficsim} or combinations of multiple objectives may also be considered.
 
\subsection{Generating Optimal Trajectories}
In practice, we do not have access to the ground truth distribution $p(y)$. To address this, we assume that $p(y) \approx q(y)$.
By making this assumption, we may insert Eq. \ref{eq:weighted_delta_distribution} into Eq. \ref{eq:definition_optimal} and obtain
\begin{equation}
    \hat{y}  \approx 
    \underset{\tilde{y}}{\operatorname{argmin}} ~
     \sum_{m=1}^M\sum_{n=1}^N \frac{w_n^m}{M}
    \text{minADE}_k(y_n^m, \tilde{y}).
    \label{eq:optimal_trajectories}
\end{equation}
Above  we replaced the risk $\mathcal{L}$ with $\text{minADE}_k$. 
The $\text{minADE}_k$ metric is differentiable with respect to $\tilde{y}$, making it compatible with gradient-based optimizers such as Adam \cite{Kingma2015_adam}.
Consequently, we can optimize the objective in Eq. \ref{eq:optimal_trajectories} with respect to the optimal prediction set with any modern automatic differentiation library. 
We refer to our method as  model-based risk minimization as it minimizes the risk under learner distribution $q(y)$ instead of the true $p(y)$ one.
We provide pseudo-code in Alg. \ref{alg:main}.

\begin{algorithm}[h]
\caption{Sampling Trajectories through Model-Based Risk Minimization}\label{alg:main}
\begin{algorithmic}
\State {\bf Inputs:} $q(y)= \sum_{m=1}^M \sum_{n=1}^N \frac{w_n^m}{M} \delta(y_n^m -y) $  \Comment{Density}
\State {\bf Outputs:} $\hat{y}=\{ \hat{y}_s \}_{s=1}^S$ \Comment{Set of optimal trajectories}
\\
\State $\hat{y} \gets$ \texttt{random\_init}$(S)$ \Comment{Randomly initialize $\hat{y}$}
\State \texttt{opt} $\gets$ \texttt{optimizer}$(\hat{y})$ \Comment{Initialize optimizer}

\While{ not converged}
    \State \texttt{loss} $\gets \sum_{m=1}^M \sum_{n=1}^N  \frac{w_n^m}{M} \text{minADE}_k(y_n^m -\hat{y})$ %\Comment{Calculate risk}
    \State \texttt{loss.backward()} 
    \State \texttt{opt.step()}  %\Comment{Update $\hat{y}$}
\EndWhile
\State \Return $\hat{y}$

\end{algorithmic}
\end{algorithm}

%% file: sections/experiments.tex
\section{Experimental Evaluations}
\label{sec:results}
Our paper introduces a novel method for generating trajectories from an ensemble \cite{Lakshminarayanan17_ensembles} of trajectory prediction models. 
The experimental section is divided into three parts.
First, we demonstrate that our method improves prediction performance when more networks are added to the ensemble. 
This is in contrast to traditional methods like selecting the most likely trajectories.
In the second experiment, we fix the ensemble of neural networks and compare our method against multiple alternatives.
Finally, we study the influence of the composition of the ensemble.  
More specifically, we study whether a mixed ensemble, an ensemble of models from the same model family, or dropout \cite{gal16_dropout} is preferable.

We benchmark our trajectory generation method on the nuScenes dataset~\cite{nuscenes}, a commonly used real-world trajectory prediction dataset for autonomous driving. 
Although many papers and models are benchmarked on the nuScenes dataset, only a few have open-source implementations available.
Throughout our experiments, we rely on LAformer~\cite{liu2024_laformer}, PGP~\cite{deo2022_pgp}, and LaPred~\cite{kim2021_lapred} to construct different ensembles of trajectory prediction models. 
These three models are among the best-performing models with open-source implementations available.
Notably, only UniTraj~\cite{feng2024unitraj} outperforms these three baselines.
However, UniTraj is trained on multiple datasets (nuScenes, WOMD~\cite{ettinger2021_womd}, and Argoverse2~\cite{wilson2021_argoverse2}), making a direct comparison unfair.

Throughout our experiments, we generate the \textit{optimal} trajectories using Alg.~\ref{alg:main} with the Adam optimizer, a learning rate of 0.1, and a fixed number of 256 steps.

\subsection{More Trajectory Proposals = Lower Prediction Error?}
\label{subsec:results_num_proposals}
\begin{figure}
\begin{center}
\includegraphics[width=1\linewidth]{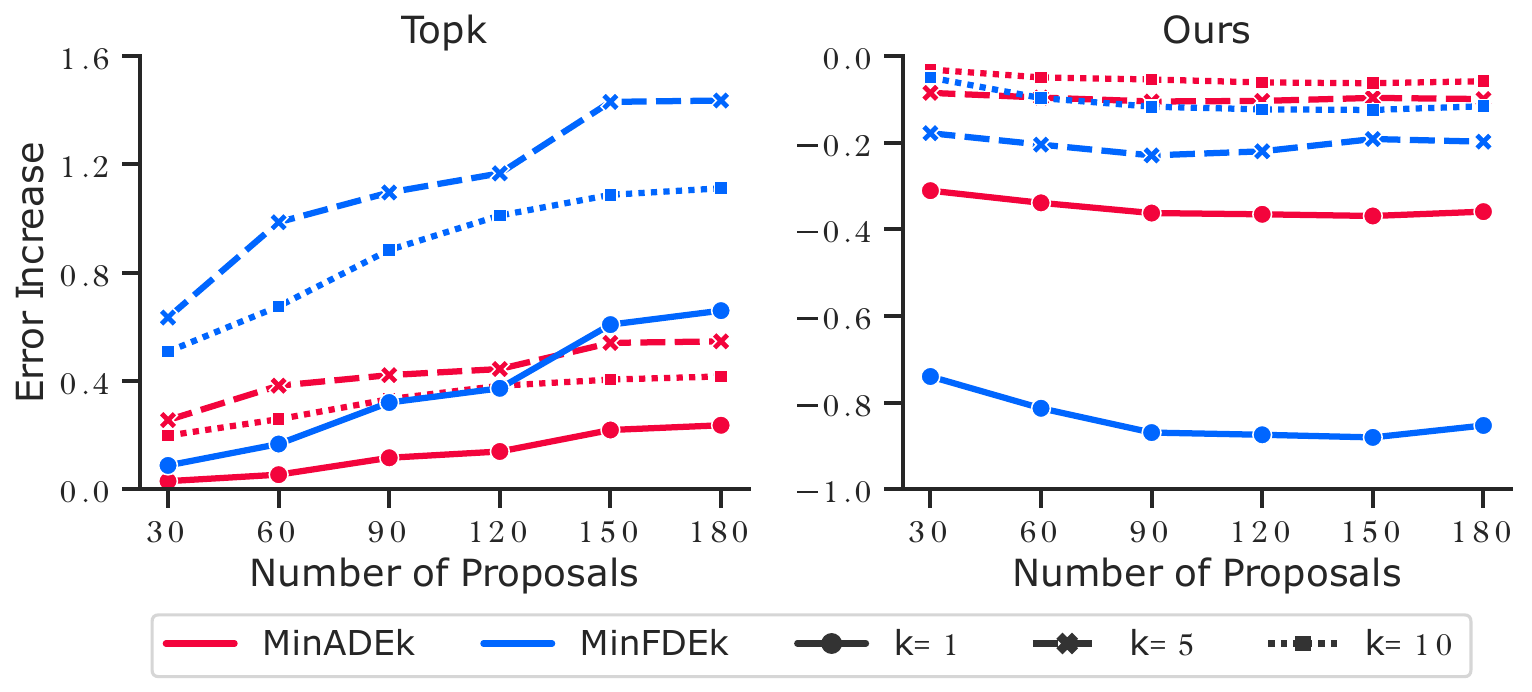}
\end{center}
\caption{Change in the prediction error as a function of the number of trajectory proposals w.r.t. PGP base performance. Trajectory proposals are equally generated from  LaPred \cite{kim2021_lapred}, LAformer \cite{liu2024_laformer}, PGP \cite{deo2022_pgp}. E.g.: 90 proposals consist of 30 proposals from each of the three models.
\textbf{Left:} Using Topk sampling strategy, which is selecting the $k$ most likely trajectories. \textbf{Right:} Our proposed sampling strategy.}
\label{fig:increasing_num_proposals}
\end{figure}

Increasing the number of trajectory proposals, and hence the number of models should intuitively decrease the $\text{\ac{minADE}}_k$ error, as we are able to cover a wider range of potential future outcomes. 
However, as the number of trajectory proposals increases, it also becomes increasingly difficult to draw the \textit{optimal} trajectories. 
We demonstrate this behavior in Fig.~\ref{fig:increasing_num_proposals}.

In this experiment, we vary the number of trajectory proposals from 30 to 180 in steps of 30. 
The proposals come equally from three different models: PGP, LaPred, and LAformer. Using the commonly used sampling technique Top-k, which selects the $k$ most likely proposals, i.e., the $k$ highest weights $w_n^m$ and their corresponding trajectories, the $\text{\ac{minADE}}_k$ and $\text{\ac{minFDE}}_k$ consistently increase as the number of proposals increases. 
This indicates that we cannot fully trust the predicted weights.
This behavior can be explained by two effects. 
Suppose we have three models, each predicting 10 trajectory-weight pairs. 
First, if all three models predict "going straight with constant velocity" as the most likely outcome, then picking the Top 3 trajectories would result in selecting the same trajectory three times, namely "going straight with constant velocity." 
This results in poor coverage of potential future outcomes and consequently high $\text{\ac{minADE}}_k$.
Additionally, if a model is uncertain about its predictions, its scores are reduced after normalization, preventing these predictions from being included in the Topk selections of the ensemble. Consequently, models with less uncertainty tend to dominate the predictions.
Contrarily, we observe the exact opposite behavior with our proposed sampling technique. As the number of trajectory proposals increases, we observe a decreasing trend for $\text{\ac{minADE}}_k$ and $\text{\ac{minFDE}}_k$.
Our sampling method considers all the predicted trajectory-weight pairs, essentially enabling communication between the models and agreeing on the \textit{optimal} predictions. 
%Our results suggest that as we increase the number of proposals, $q(y)$ moves closer to $p(y)$, making the empirical risk minimization more accurate.

\subsection{Benchmarking Against Alternative Sampling Methods}
There are many methods to sample the \textit{optimal} trajectories, raising the question of whether our proposed method is indeed the best choice. 
To address this, we benchmark our proposed sampling method against several alternatives:

\noindent \textit{(i) Uniform}: Sample uniformly $k$ trajectories.

\noindent  \textit{(ii) Categorical}: Sample $k$ trajectories from the categorical distribution with weights $w_n^m / M$.

\noindent \textit{(iii) Topk}: Select $k$ most likely trajectories as in \cite{liu2024_laformer}. 

\noindent \textit{(iv) KMeans}: Applying the KMeans algorithm to form $k$ clusters, and then selecting the closest trajectory to each cluster center from the candidate trajectories.

\noindent \textit{(v) NMS+KMeans}: Similar to \textit{(iv)} but begins with initializing clusters using \ac{NMS}, which first selects the trajectory with the highest probability and removes it from the set. Next, trajectories with an \ac{ADE} below a threshold relative to the selected trajectory are removed. This process is repeated until $k$ trajectories are selected. This technique was used in \cite{seff2023motionlm}.

In this experiment, we use a fixed set of trajectory prediction models.
More specifically, we use a collection of 12 models, with 4 models each from PGP, LaPred, and LAformer. 
This collection provides a total of 120 proposed trajectories, from which we aim to generate the \textit{optimal} samples.
Additionally, we compare our method against using each model individually, without being part of a larger model collection. 
We report both our reproduced results (the best out of 3 runs), obtained by training the models within their official open-source implementations, and results from the official publications. 
Overall, we successfully reproduce the reported results from the literature, except for LAformer for $k=5$. 
This discrepancy likely arises because we trained all models to predict 10 modes, while LAformer was specifically trained to predict 5 modes for computing $\text{\ac{minADE}}_5$.
The comparison results are presented in Tab. \ref{tab:benchmark_sampling_methods}.

\begin{table}[h]
\caption{$\text{\ac{minADE}}_k$ and $\text{\ac{minFDE}}_k$ for different $k$ values and  sampling strategies.
 LP = LaPred \cite{kim2021_lapred}, LF = LAformer \cite{liu2024_laformer}, PGP \cite{deo2022_pgp}.}
\label{tab:benchmark_sampling_methods}
\begin{center}
\resizebox{\columnwidth}{!}{
\begin{tabular}{ll*{3}{c}*{3}{c}}
\toprule
\multirow{2}{*}{Model(s)} & \multirow{2}{*}{Method} & \multicolumn{3}{c}{$\text{\ac{minADE}}_k$ ($\downarrow$)} & \multicolumn{3}{c}{$\text{\ac{minFDE}}_k$ ($\downarrow$)} \\
\cmidrule(lr){3-5} \cmidrule(lr){6-8}
& & $k=1$ & $k=5$ & $k=10$ & $k=1$ & $k=5$ & $k=10$ \\
\midrule
\multirow{6}{*}{$4\!\!\times\!\!\{\text{LP, LF, PGP}\}$} & Uniform & 5.16 & 1.79 & 1.22 & 12.15 & 3.83 & 2.28 \\
& Categorical & 3.30 & 1.70 & 1.26 & 7.74 & 3.57 & 2.41 \\
& Topk & 3.31 & 1.73 & 1.34 & 7.75 & 3.66 & 2.58 \\
& KMeans & 3.26 & 1.37 & 1.03 & 7.57 & 2.76 & 1.82 \\
& NMS+KMeans & \textbf{2.81} & \textbf{1.18} & 0.93 & \textbf{6.50} & \textbf{2.27} & 1.51 \\
& Ours & \textbf{2.81} & \textbf{1.18} & \textbf{0.89} & \textbf{6.50} & \textbf{2.27} & \textbf{1.45} \\
\midrule
LP (reproduced)  & Topk & 3.56 & 1.53 & 1.17 & 8.48 & 3.13 & 2.18 \\
LF (reproduced)  & Topk & 3.07 & 1.51 & 0.95 & 7.12 & 3.06 & 1.61 \\
PGP (reproduced) & Topk & 3.17 & 1.28 & 0.96 & 7.38 & 2.49 & 1.57 \\
LP (official)    & Topk & 3.51 & 1.53 & 1.12 & 8.12 & 3.37 & 2.39 \\
LF (official)    & Topk & NA   & 1.19 & 0.93 & NA   & NA   & NA \\
PGP (official)   & Topk & NA   & 1.30 & 1.00 & NA   & NA   & NA \\
\bottomrule
\end{tabular}
}
\end{center}
\end{table}

Comparing our proposed sampling method against various alternatives, we observe that our method consistently achieves equal or superior performance.
Among the alternatives, NMS+KMeans emerges as the strongest competitor. Our method demonstrates similar prediction performance for $k=1$ and $k=5$ and outperforms NMS+KMeans for $k=10$. 
However, we found in preliminary NMS+KMeans to be sensitive to its hyperparameters, specifically the threshold used during NMS. 
In terms of computational complexity, our method shows favorable behavior in comparison to NMS+KMeans. 
Let $S$ be the number of total proposal trajectories and $T$ the number of optimizer steps. 
Our method has a complexity of $\mathcal{O}(SkT)$, which is equivalent to the runtime of KMeans $\mathcal{O}(SkT)$.
In contrast, the runtime of NMS can vary significantly. 
In the best case, where trajectories are perfectly overlapping, NMS has a runtime of $\mathcal{O}(S)$, as a single step is sufficient. 
In the worst case, where trajectories are perfectly disconnected, NMS has a runtime of $\mathcal{O}(S^2)$, because it discards only one trajectory per step, resulting in a total of $S(S-1)/2 = \mathcal{O}(S^2)$ steps. 
Thus, the worst-case runtime of NMS+KMeans is $\mathcal{O}(SkT + S^2)$.
The alternative methods Uniform, Categorical, and Topk lag behind both KMeans and NMS+KMeans. 
This is likely because these methods do not exchange information among different models in the ensemble, leading to systematic errors as discussed in Sec. \ref{subsec:results_num_proposals}. 
This results in selecting the same trajectory multiple times if different models agree on the most likely outcome.
Next, we compare our proposed sampling method for a fixed ensemble of prediction models against using each model individually. Note that our ensemble consists only of reproduced models, as we do not have access to the original weights reported in the corresponding papers. We find that our method is able to outperform all baseline models using their default sampling technique, Top-k.
This demonstrates the advantage of using diverse models and our sampling method. 
By effectively combining information from multiple models and ensuring a more diverse set of trajectories, our proposed method mitigates the risk of redundancy and captures a broader range of potential outcomes.

\subsection{Influence of the Ensemble Composition}
We observed in the previous experiment that our composition consisting of 12 models is able to outperform each individual model. 
This raises the question of whether we can further improve performance by fine-tuning the composition of our model selection. 
In this experiment, we study the influence of different model compositions and also explore dropout \cite{gal16_dropout} as an alternative to deep ensembles \cite{Lakshminarayanan17_ensembles}. 
We report the results of this experiment in Tab. \ref{tab:benchmark_compositon}.

\begin{table}[h]
\caption{$\text{\ac{minADE}}_k$ and $\text{\ac{minFDE}}_k$ for different $k$ values and ensembling strategies using our proposed sampling strategy.  LP = LaPred \cite{kim2021_lapred}, LF = LAformer \cite{liu2024_laformer}, PGP \cite{deo2022_pgp}.}
\label{tab:benchmark_compositon}
\begin{center}
\resizebox{\columnwidth}{!}{
\begin{tabular}{lcccccc}
\toprule
{Model(s)} & \multicolumn{3}{c}{$\text{\ac{minADE}}_k$ ($\downarrow$)} & \multicolumn{3}{c}{$\text{\ac{minFDE}}_k$ ($\downarrow$)} \\
\cmidrule(lr){2-4} \cmidrule(lr){5-7}
 & $k=1$ & $k=5$ & $k=10$ & $k=1$ & $k=5$ & $k=10$ \\
\midrule
%$4 \times \{\text{LP, LF, PGP}\}$ & 120 & \textbf{2.81} & \textbf{1.18} & \textbf{0.89} & \textbf{6.50} & \textbf{2.27} & \textbf{1.45} \\
$1\! \times \! \{\text{LP, LF, PGP}\}$  & 2.89 & 1.22 & 0.93 & 6.70 & 2.37 & 1.54 \\
$3\! \times \! \text{LP}$  & 3.08 & 1.34 & 1.08 & 7.16 & 2.66 & 1.88 \\
$3\! \times \! \text{LF}$  & 2.82 & 1.20 & 0.92 & 6.52 & 2.34 & 1.54 \\
$3\! \times \! \text{PGP}$  & 2.97 & 1.22 & 0.94 & 6.87 & 2.37 & 1.55 \\
Dropout + LP & 3.15 & 1.39 & 1.14 & 7.31 & 2.78 & 2.01 \\
Dropout + LF  & 2.94 & 1.28 & 1.00 & 6.86 & 2.53 & 1.72 \\
Dropout + PGP  & 3.07 & 1.26 & 0.97 & 7.09 & 2.46 & 1.64 \\
\bottomrule
\end{tabular}
}
\end{center}
\end{table}

First, we compare the deep ensemble approach against dropout alternatives. 
For instance, we compare 3$\times$LAformer (LF) models in a deep ensemble with a single LAformer model using dropout. 
We observe that the deep ensembles consistently outperform the dropout alternatives across all models and metrics. 
This is likely due to the greater diversity offered by the deep ensemble.
In a deep ensemble, independently trained models can vary significantly, often predicting different modes of data distribution. 
In contrast, dropout within a single model does not generate such diverse predictions, typically producing variations around the same mode.
We then compare different deep ensemble compositions while keeping the number of models within each ensemble constant. 
We find that both a mixed ensemble and ensembles consisting of either 3 LAformer models or 3 PGP models show minimal differences (less than 0.03) for both $\text{\ac{minADE}}_k$ and $\text{\ac{minFDE}}_k$ at $k=1$ and $k=10$.
In contrast, an ensemble consisting solely of 3 LaPred models exhibits a larger difference of 0.30 in $\text{\ac{minFDE}}_k$ at $k=10$.
These results indicate that the performance of ensembles can remain consistent across different compositions, as long as they include strong individual prediction models. 
However, the quality of the individual models is crucial: ensembles composed of strong models, such as LAformer and PGP, exhibit better prediction results. 

%% file: sections/conclusion.tex
\section{Conclusion}
\label{sec:conclusion}
This paper presents a novel trajectory sampling technique that leverages ensemble methods for trajectory prediction. 
Our results highlight the effectiveness of using ensembles, demonstrating that our method consistently matches or surpasses the performance of existing alternatives in terms of predictive accuracy. However, one limitation of this approach is the increased memory usage and computational complexity that comes with deep ensembles.

Future work could explore variational inference methods \cite{look2023probabilisticdeepstatespace, graves11_vi}, which could potentially reduce the overhead of deep ensembles while maintaining high-quality predictions, making the technique more applicable in resource-constrained environments. Additionally, we propose further research into optimizing other targets beyond $\text{\ac{minADE}}_k$. 
By investigating optimization criteria such as combinations of $\text{\ac{minADE}}_k$, offroad rate, and infraction rate, the method could be refined to better suit specific real-world applications where safety and adherence to constraints are critical.